\documentclass[journal]{IEEEtran}
\usepackage{url}
\usepackage[numbers]{natbib}
\usepackage[bookmarks=true]{hyperref}
\usepackage{amsmath,mathtools,amsfonts,amssymb,graphicx}
\usepackage{algorithm}
\usepackage[noend]{algpseudocode}
\usepackage{float,color,caption,wrapfig}
\usepackage[mathscr]{euscript}
\DeclareMathOperator*{\argmin}{\arg\!\min}
\DeclareMathOperator{\sign}{sign}

\newcommand\NoDo{\renewcommand\algorithmicdo{}}
\newcommand\NoThen{\renewcommand\algorithmicthen{}}

\begin{document}

\title{Algorithmic Design for Embodied Intelligence\\ in Synthetic Cells}

\author{Ana~Pervan, Todd~D.~Murphey
\thanks{A.~Pervan and T.~D.~Murphey are with the Department
of Mechanical Engineering, Northwestern University, Evanston,
IL, 60208 USA. e-mail: anapervan@u.northwestern.edu}
\thanks{Manuscript received June 24, 2019; revised May 27, 2020.}}



\maketitle

\begin{abstract}
In nature, biological organisms jointly evolve both their morphology and their neurological capabilities to improve their chances for survival. Consequently, task information is encoded in both their brains and their bodies. In robotics, the development of complex control and planning algorithms often bears sole responsibility for improving task performance. This dependence on centralized control can be problematic for systems with computational limitations, such as mechanical systems and robots on the microscale. In these cases we need to be able to offload complex computation onto the physical morphology of the system. To this end, we introduce a methodology for algorithmically arranging sensing and actuation components into a robot design while maintaining a low level of design complexity (quantified using a measure of graph entropy), and a high level of task embodiment (evaluated by analyzing the Kullback-Leibler divergence between physical executions of the robot and those of an idealized system). This approach computes an idealized, unconstrained control policy which is projected onto a limited selection of sensors and actuators in a given library, resulting in intelligence that is distributed away from a central processor and instead embodied in the physical body of a robot. The method is demonstrated by computationally optimizing a simulated synthetic cell.
\end{abstract}

Note to Practitioners:
\begin{abstract}
As robotic systems approach the micron scale, designing them to be fully autonomous will rely less on on-board computation and more on component selection and design.  In this paper we are motivated by synthetic cells---microscopic devices with limited actuation, sensing, and memory components. We apply tools from optimal control, graph theory, and information theory to develop a methodology for designing the electronic circuitry that relates actuation to sensing using memory and physically-realizable transformations (e.g., simple logical operators).  Results indicate that encoding task information in the physical body of a robot via a simple control policy leads to successful task performance.  In future work, we plan to apply these methods to different robotic systems and to experimentally employ these designs on actual synthetic cells.
\end{abstract}

\begin{IEEEkeywords}
Design methodology, morphological operations, information theory
\end{IEEEkeywords}

\IEEEpeerreviewmaketitle

\section{Introduction}
\IEEEPARstart{E}{mbodied} intelligence, the coupling of a system's controller and morphology, has been studied for quite some time \cite{brooks1991}, \cite{pfeifer2007book}, \cite{pfeifer2007science}, often in the context of biologically inspired systems \cite{dickinson2000}, \cite{spagna2007}, \cite{wood2008}. Recently, some significant and comprehensive efforts have been made toward implementing components of embodiment in robotic applications \cite{banarse2019}, \cite{howard2019}. But still, most robot designers opt for approaches using centralized computations to manipulate existing robotic platforms, rather than offloading some of the computational effort onto a robot's morphology. In this work, we are motivated by a system that is computationally limited but flexible in terms of physical design, and is therefore an ideal candidate to take advantage of embodied intelligence.

\begin{figure}
\centering
\includegraphics[width=7cm]{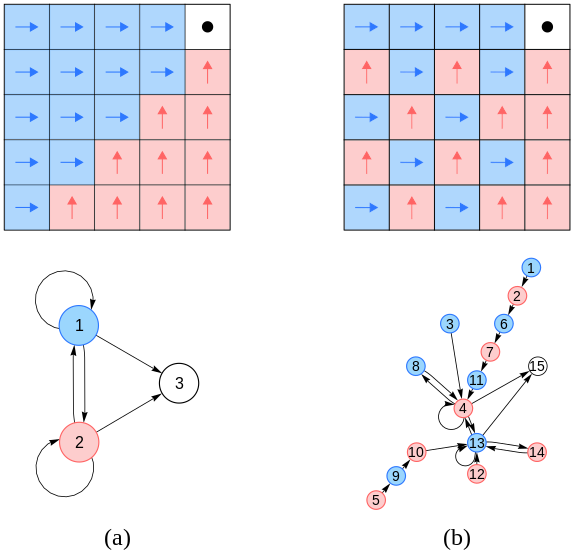}
\caption{A simple example of a control policy. Here the state space is a two dimensional $5 \times 5$ grid and the desired state is in the upper right of the state space. At each state, the suitable control is indicated by its assigned color. (a) A \emph{simple} control policy with only three different states in the finite state machine. (b) A \emph{complex} control policy, with fifteen different states in the finite state machine. Both achieve the task with different implementations.}
\label{fig-policy}
\end{figure}

The main challenge with embedding control information in material properties is the contradiction between continuous, often complex, classical control and the discrete, simpler capabilities that a materials-based system is likely to have. A conflict exists between equipping a robot with what is sufficient and what is necessary---what can enable a robot to succeed (and is likely complex) and what is minimally required for it to achieve its goal (and is necessarily simple). 

This is especially evident when designing for robots without any on-board, CPU-based, traditional computational capabilities. Some robotic systems employ embodied intelligence (which we'll also refer to as embodied computation) to reduce weight and energy---for example fully mechanical devices, like passive dynamic walkers \cite{collins2005bipedal}, \cite{tedrake2004walker} or those used in prosthetic limbs \cite{arelekatti201prosthetic}---while others must \emph{necessarily} resort to embodied computation because of scale, for example, robots on the micro- or nano-scales  \cite{esteban2018nano}, \cite{li2018nano}.

We first examine the relationship between control policy design and physical robot body design. A control policy assigns an action for a system at each time or state. Symbolic control policies have been useful in robot control and motion planning \cite{belta2007symbolic}, \cite{mavrommati2016cap} for systems with limited computational power \cite{pola}. An example is shown in Fig.~\ref{fig-policy}, where the system consists of three possible control modes: move right (blue), move up (red), and stay still (white) and its goal is to navigate to the upper right corner of the grid world. A simple robot placed in this environment with one of these policies in its memory would be able to achieve its task without any traditional computation, replacing logical operators such as inequalities with physical comparators to relate sensor states to control actions. 

Figure~\ref{fig-policy} illustrates the difference in complexity between these two policies. The policy in Fig.~\ref{fig-policy}~(a) requires a combination of sensors that are capable of differentiating between three different regions of the state space, and the policy in Fig.~\ref{fig-policy}~(b) requires sensors that are able to discern between fifteen different regions. It is simpler to physically implement the robot design implied by the control policy in Fig.~\ref{fig-policy}~(a) than the policy in Fig.~\ref{fig-policy}~(b).

We pose the physical design problem as the projection of a policy onto an admissible set of physical sensor-actuator interconnections, the complexity of which must be managed during policy iteration. This complexity is a measure of logical interconnections between sensor states and control modes (which correspond to arrows on the graphs in Fig.~\ref{fig-policy}). The policy projection will be performed both by computing a control policy assuming discrete control modes (and continuous sensing) and then projecting onto a discrete sensor set, and by generating a control policy assuming discrete sensing (and unconstrained control authority) and then projecting onto a set of discrete control modes.

After reviewing related work in Section~\ref{sec-work}, this methodology will be explored in terms of an extended example. The example system, called a synthetic cell, will be introduced in Section~\ref{sec-problem}. The primary contributions of this work can be summarized as follows: 
\begin{enumerate}
\item Quantitative definitions are given for design complexity and task embodiment, described in Section~\ref{sec-complexity}.
\item An iterative algorithm is developed in Section~\ref{sec-controls-first}, and is used to create control policies with low design complexity while increasing task information.
\item A projection operator is presented in Section~\ref{sec-control-proj}, which projects a low complexity control policy onto a physically realizable set of sensor-actuator interconnections.
\item A methodology for algorithmically organizing components for robot design is established. The procedure begins either with a control policy based on a discrete set of actuators (Sec.~\ref{sec-controls-first}) and interconnects them with different possible sets of sensors (Sec.~\ref{sec-control-proj}) or begins with a control policy generated with a discrete set of sensors (Sec.~\ref{sec-sensing-first}) and combines them with a selection of discrete actuators (Sec.~\ref{sec-sensing-proj}).
\end{enumerate}

These are supported by simulations of synthetic cells. A version of this work was published in a conference paper \cite{pervan2018wafr}.

\section{Related Work} \label{sec-work}

Policy Optimization while Evolving Morphology (POEM) \cite{banarse2019}, evolves the physical body of a continuously controlled reinforcement learning agent and analyzes the relative importance of body changes using cooperative game theory.  The POEM method was shown to produce stronger agents than optimizing the control policy alone. A common motivation of the work in \cite{banarse2019} and this paper is the theory that a physical body that is well-suited to a task is easier, and simpler, to control (and, in \cite{banarse2019}, easier to \emph{learn} to control). In this work, we characterize design updates in terms of moving task information from centralized computations in control calculations to embodied computation in the physical body.

The work in \cite{censi2015codesign}, \cite{censi2016cyclic}, and \cite{censi2017uncertainty} defines design problems as relations between functionality, resources, and implementation and shows that despite being non-convex, non-differentiable, and noncontinuous, it is possible to create languages and optimization tools to define and automatically solve design problems. The optimal solution to a design problem is defined as the solution that is minimal in resources usage, but provides maximum functionality. We apply this definition by proposing a min-max problem in which the goal is to minimize design complexity (representative of the amount of sensors and actuators required, i.e., the resources), and maximize task embodiment (i.e., the functionality of the design).

A method for automatically designing action-based sensors was explored in \cite{erdmann1995}. This was done by generating a strategy for a robot task using a planner that assumes perfect sensing, and using that plan to specify sensors that tell the robot where to execute each action. The methodology in \cite{erdmann1995} is very similar to the work presented in this paper, which also first develops a control policy assuming perfect sensing (or perfect actuation) and then specifies discrete sensors (or actuators) that approximate the original strategy. This paper differs from the contributions in \cite{erdmann1995} by taking complexity into account and by also designing actuators.

Robotic primitives are introduced in \cite{okane2008power} as independent components that may involve sensing or motion, or both. These are implemented in this work as actuator and sensor libraries from which we allow our algorithm to choose components. Task embodiment, which is defined in Sec.~\ref{sec-complexity}, parallels the dominance relation proposed in \cite{okane2008power} that compares robot systems such that some robots are stronger than others based on a sensor-centered theory of information spaces. 

Similarly, our definition of design complexity (Sec.~\ref{sec-complexity}) parallels an existing notion of conciseness, presented in \cite{okane2017concise}. The results in \cite{okane2017concise} are motivated by circumstances with severe computational limits, specifically addressing the question of how to produce filters and plans that are maximally concise subject to correctness for a given task. This is very related to our goal of finding the simplest way to physically organize sensors and actuators so that a (computationally limited) robot can achieve a given task.

The work presented in \cite{karaman2011sampling} produces asymptotically optimal sampling-based methods and proposes scaling laws to ensure low algorithmic complexity for computational efficiency. These algorithms were originally developed for path planning, but we apply similar ideas for generating simple control policies. The methods described in \cite{karaman2011sampling} start with an optimal, infinite complexity solution, and from that develop simpler plans. In Sec.~\ref{sec-controls-first}, we start with a zero complexity policy and move towards more complex, better performing solutions---while maintaining a level of computational complexity appropriate for physical implementations of embodied computation.

\section{Motivating Example: The Synthetic Cell} \label{sec-problem}
How can we use control principles to organize sensor components, actuator components, and their interconnections to create desired autonomous behavior, without relying on traditional computation? To answer this question we will consider the extended example of a synthetic cell---a small robot that only has a finite number of possible sensor and actuator states and potential pairings between them \cite{janus}. The purpose of this example system is to show a concrete implementation of the methods in Sections~\ref{sec-controls-first}-\ref{sec-sensing-proj}, and to illustrate the relationship between control policy design and physical robot body design.

A synthetic cell is a mechanically designed microscopic device with limited sensing, control, and computational abilities \cite{janus}; it is essentially an engineered cell. A synthetic cell exists in a chemical bath and generates movement by interacting with its environment using chemical inhibitors, and it contains simple circuits that include minimal sensors and very limited nonvolatile memory \cite{strano}. Such a device is 100$\mu$m in size or less, rendering classical computation using a CPU impossible. But these simple movement, sensory, and memory elements can be combined with a series of physically realizable logical operators to enable a specific task.

For the example in this paper, a synthetic cell operates in a two dimensional space, and its control authority is the ability to be attracted toward a specific chemical potential. So at any location $(x,y)$ the robot may choose a control mode $\sigma \in \left\{\sigma_{0}, \sigma_{1}, \sigma_{2}, \sigma_{3}, \sigma_{4}, \sigma_{5}, \sigma_{6} \right\}$, where $\sigma_{0}$ is zero control, and the other six modes are a potential that the synthetic cell can be attracted to (their locations are shown in Fig.~\ref{fig-control-proj}), with dynamics

\vspace{-2mm}
\begin{equation}
\textbf{x} = 
\begin{pmatrix}
x\\
\dot{x}\\
y\\
\dot{y}
\end{pmatrix},
 \quad
f(\textbf{x},u) = 
\begin{pmatrix}
\dot{x}\\
\frac{\sign{(x_{S_n}-x})}{r_{n}^{2}}\\
\dot{y}\\
\frac{\sign{(y_{S_n}-y})}{r_{n}^{2}}
\end{pmatrix},
\end{equation}
\noindent 

where $r_{n}$ is the distance from the synthetic cell to source $n$ and $(x_{S_n},y_{S_n})$ are coordinates of the source locations. Because of the inverse squared terms in the dynamics, chemical sources that are nearer to the synthetic cell will be able to accelerate the cell faster than those that are far away. Boundary conditions, like those discussed in \cite{burgess2017}, are necessary to avoid an infinite acceleration as $r_n \rightarrow 0$. We included a very small boundary $\varepsilon$ around the chemical source \cite{burgess2017}, so that the cell cannot be co-located with the source. The maximum velocity was also bounded, to mimic terminal velocity in a fluid.

The control synthesis problem is to schedule $\sigma$ in space $(x,y)$, based on an objective (in this case, to approach a point $\mathcal{P}$) specified in a cost function $J$ \eqref{eq-cost} made up of a running cost $\ell(x(t),u(t))$ \eqref{eq-runningcost} and a terminal cost $m(x(t_f))$ \eqref{eq-terminalcost}.

\vspace{-3mm}
\begin{equation} \label{eq-cost}
J(x(t),u(t)) = \int_{0}^{t_f} \ell(x(t),u(t))dt + m(x(t_f)).
\end{equation}

\vspace{-2mm}
\begin{equation} \label{eq-runningcost}
\ell(x,u) = (x-x_d)^{T} Q(x-x_d) +  u^{T} R u
\end{equation}

\vspace{-2mm}
\begin{equation} \label{eq-terminalcost}
m(x) = (x-x_d)^{T} P_1 (x-x_d). 
\end{equation}

For our simulations, we used the parameters: prediction time horizon $T = 0.1 s$; time step $t_s = 0.02 s$; final time $t_f = 5 s$; desired state\footnote{Point $\mathcal{P}$ is slightly off center, to avoid adverse effects of symmetry. This is reflected in the asymmetry of the resulting control policies (e.g., green and orange not being perfectly even).} $x_d = [2-\frac{\pi}{20},4-\frac{\pi}{15},0,0]^T$; size of the source $\varepsilon = 0.001$; maximum velocity $v_{max} = 0.4$; cost weights $Q = P_1 = diag[10,10,0.001,0.001]$ and $R=0$; and source locations $(x_{S_1}, y_{S_1}) = (1,5)$, $(x_{S_2}, y_{S_2}) = (3,5)$, $(x_{S_3}, y_{S_3}) = (1,3)$, $(x_{S_4}, y_{S_4}) = (3,3)$, $(x_{S_5}, y_{S_5}) = (1,1)$, and $(x_{S_6}, y_{S_6}) = (3,1)$. The state space, desired point, and chemical sources are all shown in Fig.~\ref{fig-control-proj}~(a)

\section{Design Complexity and Task Embodiment} \label{sec-complexity}

Graph entropy \cite{anand2009entropy},~\cite{dehmer2011graphentropy} will be used as a measure of design complexity for comparing robot designs. The complexity of a control policy is equated with the measure of entropy of its resulting finite state machine.

A finite state machine consists of a finite set of states (nodes), a finite set of inputs (edges), and a transition function that defines which combinations of nodes and edges lead to which subsequent nodes \cite{ramadage1987}. The finite set of nodes that the system switches between are the \emph{control modes}, and the edges---inputs to the system which cause the control modes to change---are the \emph{state observations} (Fig.~\ref{fig-policy}~(a)).

Finite state machines and their corresponding adjacency matrices are generated numerically, by simulating a synthetic cell forward for one time step, and recording control modes assigned at the first and second states. These control mode transitions are counted and normalized into probabilities, and the resulting data-driven adjacency matrix $A$ is used in the entropy calculation,

\vspace{-2mm}
\begin{equation}
h = -\sum_i{A(i) \log \left( A(i) \right)}
\label{eq-entropy}
\end{equation}
\vspace{-2mm}

which results in a complexity measure $h$ for each robot design. \textcolor{black}{This measure of complexity is more informative than other metrics (e.g., simply counting states) because it is a function of the \emph{interconnections} between states---which is what we want to minimize in the physical design.}

We define \emph{task embodiment} as the amount of information about a task encoded in a robot's motion (not to be confused with embodiment found in human-robot interaction \cite{laviers}, \cite{shell}).  We focus on this motion-based task information so that the design update can be characterized in terms of moving task information from the centralized computations in the control calculations to embedded computation in the physical body. One measure that captures how much information one system encodes about another system is Kullback-Leibler divergence. Here we measure the K-L divergence between a distribution representing the task, $P$, and a distribution representing the robot design, $Q$ \cite{bishop2006},

\vspace{-2mm}
\begin{equation}
D_{KL} \left( P \middle\| Q \right) = -\sum_{\textbf{x}}{P(\textbf{x}) \log \left(\frac{Q(\textbf{x})}{P(\textbf{x})}\right)}.
\label{eq-KL}
\end{equation}
\vspace{-2mm}

To define the goal task distribution $P$, a model predictive controller (MPC) is used to simulate the trajectories of a robot with an ideal (centralized, unlimited in sensing (Sec.~\ref{sec-controls-first}) or actuation (Sec.~\ref{sec-sensing-first})) controller. The same method is used to generate a distribution $Q$  that represents the robot design---this time simulating trajectories using the generated control policy. Task embodiment is a measure of the difference in task executions between a robot with an ideal controller and a resource-limited robot with some embodied intelligence. We use Eq.~\eqref{eq-KL} to compare the two distributions of trajectories: a low measure of K-L divergence indicates that the distributions are similar, and implies a high level of task embodiment, and therefore a better robot design.

In other words, if a task is well-embodied by a robot, only a simple control policy is necessary to execute it. Otherwise, more information, in the form of a more complex control policy, is required. To construct these control policies, we will explore two opposing procedures: optimizing with respect to actuation assuming unconstrained sensing (Sec.~\ref{sec-controls-first}) and projecting onto discrete sensor sets (Sec.~\ref{sec-control-proj}), or optimizing with respect to sensing assuming unconstrained actuation (Sec.~\ref{sec-sensing-first}) and projecting onto discrete control modes (Sec.~\ref{sec-sensing-proj}).

\section{Control Policy Generation: Actuation First} \label{sec-controls-first}

\begin{figure}[t]
\centering
\includegraphics[width=8.5cm]{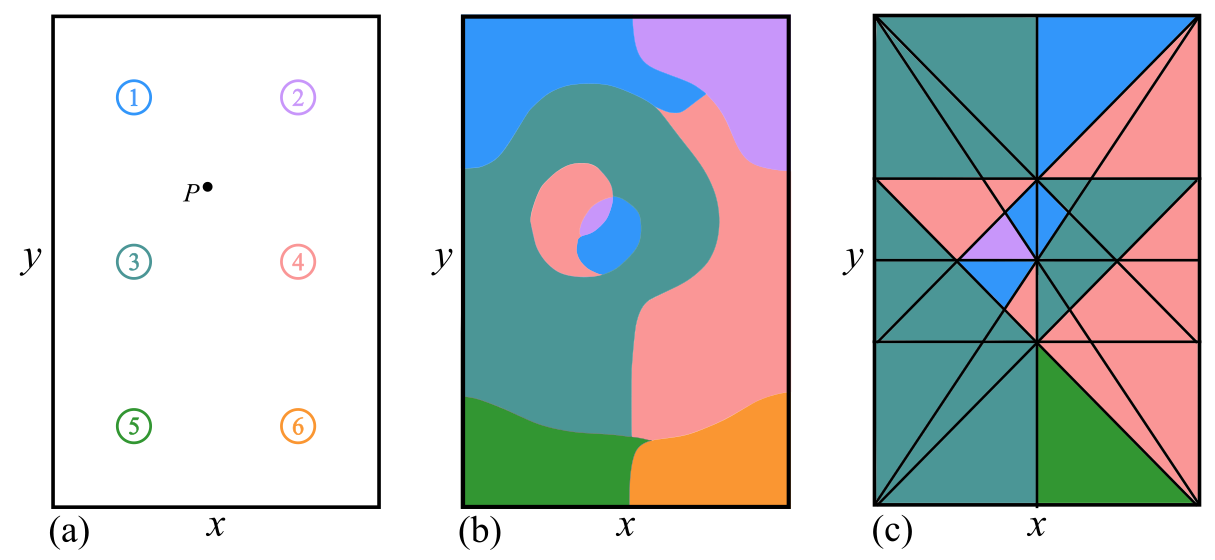}
\caption{(a) The state space and controls for the synthetic cell example system introduced in Sec.~\ref{sec-problem}. At any location in the state space, the robot is able to choose one of seven different control modes: attraction to chemical potentials at the six different sources or zero control. The goal is to reach $\mathcal{P}$. (b) A control policy for this system generated with discrete controls and unconstrained sensing. (c) The control policy projected onto a feasible set of sensor states. (These figures will be explained in detail in Sec.~\ref{sec-controls-first} and Sec.~\ref{sec-control-proj}.)}
\label{fig-control-proj}
\end{figure}

The optimization problem of minimizing implementation complexity while maximizing task embodiment is challenging, with many reasonable approaches. We use techniques from hybrid optimal control because of properties described next.

It was proven in \cite{decarlo} that optimal control of switched systems will result in a chattering solution with probability 1. Chattering is equivalent to switching control modes very quickly in time. In the case of these control policies, this translates to switching between control modes very quickly in state. As a result, an implementation of the optimal control policy would be highly complex. Instead of an optimal solution, we are looking for a ``good enough,'' near optimal solution that results in a minimal amount of mode switching. It was shown in \cite{caldwellNAHS} that the mode insertion gradient (MIG), which will be discussed in Section~\ref{sec-mig}, has useful properties, including that when the MIG is negative at a point it is also negative for a region surrounding that point, and that a solution that switches modes slowly can be nearly as optimal as a chattering solution. 

This section will first review the topics of switched systems \cite{decarlo}, \cite{caldwellNAHS}, \cite{Egerstedt03optimalcontrol} and the use of needle variations for optimization \cite{egerstedt2006transitiontime}, \cite{shaikh2003optimal}, \cite{xu2002optimal}, then develop an algorithm for building low complexity control policies. The algorithm creates a simple control policy under the assumption that the system has perfect knowledge of its state. Mapping this policy to physically realizable sensors is the subject of Section~\ref{sec-control-proj}.

\subsection{Switched Systems}
A switched-mode dynamical system is typically described by state equations of the form
\begin{equation}
\dot{x}(t) = \left\{f_{\sigma}(x(t))\right\}_{\sigma \in \Sigma}
\label{eq-dyn}
\end{equation}
\noindent 

with $n$ states $x:\mathbb{R} \to X \subseteq \mathbb{R}^{n}$, $m$ control modes $\Sigma = \left\{\sigma_{1},\sigma_{2},\ldots,\sigma_{m}\right\}$,  and continuously differentiable functions $\left\{f_{\sigma}:\mathbb{R}^{n+m} \to \mathbb{R}^{n}\right\}_{\sigma \in \Sigma}$ \cite{egerstedt2006transitiontime}. Such a system will switch between modes a finite number of times $N$ in the time interval $[0,t_f]$. The control policy for this type of switched system often consists of a mode schedule containing a sequence of the switching control modes $\mathcal{S} = \left\{\sigma(1),\ldots,\sigma(N)\right\}$ and a sequence of switching times $\mathcal{T} = \left\{\tau_{1},\ldots,\tau_{N}\right\}$ \cite{Egerstedt03optimalcontrol}, \cite{xu2002optimal}.

In this paper, we will consider a similar switched-mode system, but instead of implementing an algorithm to optimize transition \emph{times} between modes (so that control modes are scheduled as a function of time $\sigma(t)$), we optimize transition \emph{states} (so that control modes are a function of state $\sigma(x)$). This way a robot can directly map sensory measurements of state to one of a finite number of control outputs.

\subsection{Hybrid Optimal Control} \label{sec-mig}
Let $\ell : \mathbb{R}^n \to \mathbb{R}$ be a continuously differentiable cost function, and consider the total cost $J$, defined in Eq.~\eqref{eq-cost}. We use the Mode Insertion Gradient (MIG) \cite{egerstedt2006transitiontime}, \cite{shaikh2003optimal}, \cite{xu2002optimal} to optimize over the choice of control mode at every state. The MIG measures the first-order sensitivity of the cost function~\eqref{eq-cost} to the application of a control mode $\sigma_{i}$ for an infinitesimal duration $\lambda \to 0^{+}$. The MIG $d_{i}(x)$ is defined
\begin{equation}
d_{i}(x) = \frac{dJ}{d\lambda^{+}}\biggr\rvert_t = \rho(t)^{T}(f_{\sigma_{i}}(x(t))-f_{\sigma_{0}}(x(t))).
\label{eq-MIG}
\end{equation}
The adjoint variable $\rho$ is the sensitivity of the cost function
\begin{equation}
\dot{\rho} = -(\frac{\partial f_{\sigma_{0}}}{\partial x}(x(t)))^{T}\rho-(\frac{\partial \ell}{\partial x}(x(t)))^{T}, \quad \rho(t_f)=0.
\label{eq-adjoint}
\end{equation}

The derivation of these equations is discussed in \cite{egerstedt2006transitiontime}, \cite{shaikh2003optimal}, \cite{xu2002optimal}, but the key point is that $d_{i}(x)$ measures how much inserting a mode $\sigma_{i}$ locally impacts the cost $J$. When $d_{i}(x)<0$, inserting control mode $\sigma_{i}$ at state $x$ will decrease the cost throughout a volume around $x$, meaning a descent direction has been found for that state.  The MIG can be calculated for each mode so that $d(x)$ is a vector of $m$ mode insertion gradients: $d(x) = [d_{1}(x),...,d_{m}(x)]^{T}$. Therefore the best actuation mode (i.e., the mode with the direction of maximum descent) for each state $x$ has the most negative value in the vector $d(x)$.

\begin{figure}[t]
\centering
\includegraphics[width=5.3cm]{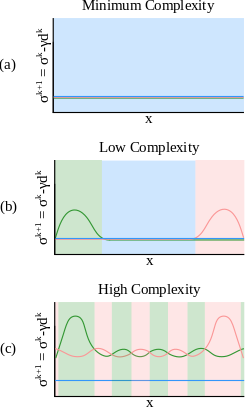}
\caption{The curves show $\sigma^{k+1}=\sigma^{k}-\gamma d^{k}$ for three different control modes $\sigma_{1}$ (blue), $\sigma_{2}$ (green), and $\sigma_{3}$ (red), where $\gamma$ is the line search parameter and the background colors indicate which mode is assigned to state $x$ in the control policy. As step size $\gamma$ increases from top to bottom, the magnitude of $\gamma d^{k}(x)$ surpasses that of the default control $\sigma^{k}(x)$ (here the default control is $\sigma_{1} =0$). (a) Minimum Complexity: Only one control mode is assigned throughout the entire state space. (b) Low Complexity: A few control modes are employed, indicating that the cost function can be reduced by including these extra control modes. (c) High Complexity: The control mode switches often but the values are similar, indicating that there is no significant difference in cost between these modes---despite the large increase in complexity (i.e., chattering).}
\label{fig-complex}
\end{figure}

As long as the dynamics $f(x(t))$ are real, bounded, differentiable with respect to state, and continuous in control and time and the incremental cost, $\ell(x(t))$, is real, bounded, and differentiable with respect to state, the MIG is continuous \cite{caldwellNAHS}. Sufficient descent of the mode insertion gradient is proven in \cite{caldwellNAHS}, where the second derivative of the mode insertion gradient is shown to be Lipschitz continuous under assumptions guaranteeing the existence and uniqueness of both $x$, the solution to the state equation Eq.~\eqref{eq-dyn}, and $\rho$, the solution of the adjoint equation Eq.~\eqref{eq-adjoint}. Combining this with the results of \cite{karaman2011sampling}, one can conclude that any sufficiently dense finite packing will also satisfy the descent direction throughout the volume of packing. As a result, although chattering policies may be the actual optimizers, finite coverings will generate descent throughout the state space, resulting in a non-optimal but ``good enough'' solution. This provides the required guarantee that we can locally control the complexity of the policy as a function of state. This will be discussed further in Sec.~\ref{sec-alg}.

\begin{figure}[t]
\centering
\includegraphics[width=0.3\textwidth]{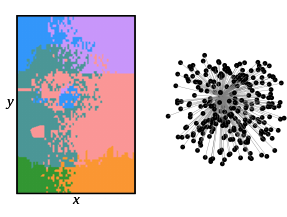}
\caption{A chattering control policy. The corresponding graph has entropy $h = 7.6035$.}
\label{fig-chatter}
\end{figure}

Figure~\ref{fig-complex} illustrates differences in complexity as a result of optimizing using the mode insertion gradient. The magnitude of the curves is the default control (the control we are comparing to) minus the step size (a scaling factor, and also the line search parameter) multiplied by the MIG. Therefore the magnitude of these plots correspond to the amount of reduction in cost that can be achieved by locally employing each control mode at the state $x$. The complex policy illustrated in Fig.~\ref{fig-complex}~(c), occurs in simulation of the chattering policy of Fig.~\ref{fig-chatter}. This happens when there is similar utility in employing more than one mode in a region---there is only marginal benefit in choosing one control mode over another, which results in increased complexity. 

\subsection{Iterative Algorithm} \label{sec-alg}
An algorithm is introduced that can reduce the complexity of a control policy in as little as one iteration, based on the work in \cite{caldwell2013projection}, \cite{caldwellNAHS}. 


\begin{algorithm}[H]
\caption{Iterative Line Search Optimization}
\label{alg-1}
\begin{algorithmic}
\State \textbf{Input Parameters:} $\epsilon_h$, $\epsilon_J$
\State \textbf{Initialize Variables:} $k = 0$, $\gamma = 0.001$
\State Choose default policy $\sigma^{0}(x)$
\State Calculate initial cost $J(\sigma^{0}(x))$
\State Calculate initial complexity $h^{0}$
\State Calculate initial descent direction $d^{0}(x)$ 
\State $h^{k-1} = \infty$
\NoDo
\While {$h^{k} < h^{k-1} + \epsilon_{h}$}
    \NoDo
	\While{$J(\sigma^{k}(x)) < J(\sigma^{k+1}(x)) +\epsilon_{J}$}
    	\State Re-simulate $\sigma^{k+1}(x) = \sigma^{k}(x)-\gamma d^{k}(x)$
		\State Compute new cost $J(\sigma^{k+1}(x))$
		\State Increment step size $\gamma$ 
	\EndWhile
	\State Calculate new complexity $h^{k+1}$
	\State Calculate $d^{k+1}(x)$
    \State $k = k + 1$
\EndWhile
\end{algorithmic}
\end{algorithm}

\begin{figure*}[ht]
\centering
\includegraphics[width=0.9\textwidth]{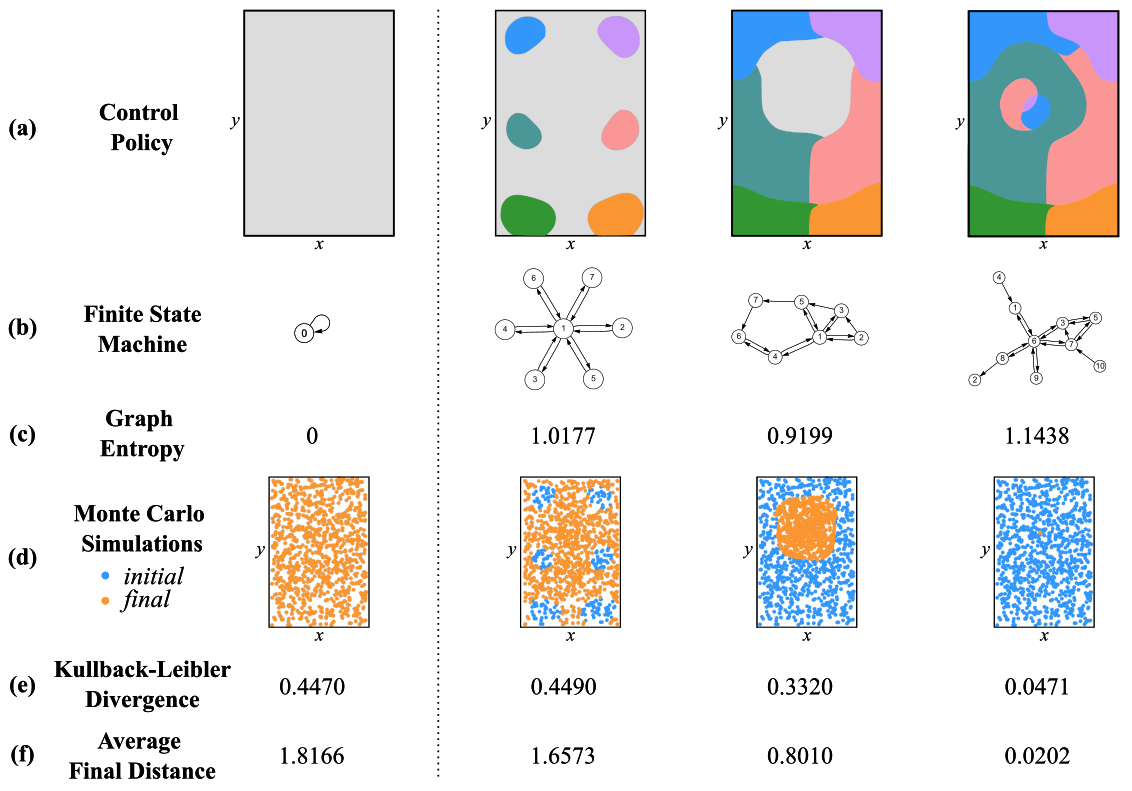}
\caption{Control policies for the system in Fig. ~\ref{fig-control-proj}, starting from a null initial policy. (a) The control policies, mapping state to control for various iterations of the line search. (b) A finite state machine representation of each policy, representative of the complexity of the system.  (c) The design complexity value calculated from Eq. \eqref{eq-entropy}.  (d) 1000 Monte Carlo simulations illustrate the results of random initial conditions using the associated control policies. (e) The Kullback-Leibler divergence between the goal task distribution and the distribution generated by the control policy, from Eq.~\eqref{eq-KL}. (f) The average final distance of the final states of (d) from the desired point $\mathcal{P}$.}
\label{fig-policies}
\end{figure*}

In this line search algorithm the default control may be chosen arbitrarily, but for simplicity we will show an example using a null default policy (in Fig.~\ref{fig-policies}).

The cost $J(\sigma^{k}(x))$ of the entire policy is approximated by simulating random initial conditions forward in time and evaluating the total cost function for time $t_f$. We use cost $J$ rather than task embodiment $D_{KL}$ as the objective function because the line search in the algorithm is a function of $d(x)=\frac{dJ}{d\lambda}$. This way, the algorithm decreases the objective function (plus tolerance $\epsilon_{J}$) each iteration. After choosing a default policy $\sigma^{0}(x)$, computing the initial cost $J(\sigma^{0}(x))$, and calculating the initial entropy $h^{0}$ (using Eq.~\eqref{eq-entropy}) the initial descent direction $d^{0}(x)$ is calculated for the set of points in $S$, as described in Sec.~\ref{sec-mig}. A line search \cite{armijo1966} is performed to find the maximum step size $\gamma$ that generates a reduction in cost in the descent direction $d^{k}(x)$, and then the policy $\sigma^{k}(x)$ is updated to the policy $\sigma^{k+1}(x)$. The new design complexity $h^{k+1}$ and descent directions $d^{k+1}(x)$ are calculated, and this is repeated until the cost can no longer be reduced without increasing the complexity beyond the threshold defined by $\epsilon_{h}$.

The tolerances $\epsilon_{h}$ and $\epsilon_{J}$ are design choices based on how much one is willing to compromise between complexity and performance. In the example illustrated in Sec.~\ref{sec-2d}, the value for $\epsilon_{h}$ is significant because it represents the allowable increase in complexity---how much complexity the designer is willing to accept for improved task embodiment. For these figures, we used $\epsilon_{h}=1.25$ and $\epsilon_{J}=10$.

This algorithm enforces low design complexity, meaning it will not result in chattering outputs. The work in \cite{caldwellNAHS} showed that if $d(x(\tau))<0$ then there exists an $\epsilon > 0$ such that $d(x(t)) < 0 \  \forall \  t \in [\tau-\epsilon,\tau+\epsilon]$. Since $d(x)$ is continuous in $x$ (as discussed in Section~\ref{sec-mig}), $d(x_{0})<0$ implies that there exists an $\epsilon > 0$ such that $d(x) < 0 \  \forall \  x \in B_{\epsilon}(x_{0})$. Note that each point in $B_{\epsilon}(x_{0})$ does not necessarily have the same mode of \emph{maximum} descent, but they do each have a \emph{common} mode of descent.

The MIG serves as a descent direction for a volume in the state space, rather than just at a point. This property allows us to assign one control mode throughout a neighborhood so that instead of choosing the optimal control mode (the direction of maximum descent) at each point and causing chattering, we select a good control mode (a direction of descent) throughout a volume and maintain relative simplicity in the policy. Figure~\ref{fig-chatter} shows a control policy that is the result of assigning the \emph{optimal} control mode at each point, which results in chattering.

\subsection{Examples} \label{sec-2d}
Figure~\ref{fig-policies} begins with an initial control policy of zero control throughout the state space, and increases design complexity and task embodiment until the line search algorithm converges to a new control policy. Monte Carlo simulations were performed with $1000$ random initial conditions, shown in row~(d) and the average distance of the final points from the desired point is shown in row~(f). Most interesting are the trends in rows~(c) and~(e). These correspond to the min-max problem posed earlier, in which we attempt to minimize design complexity, computed using graph entropy~(c), and maximize task embodiment, calculated using K-L divergence~(e). The graph entropy in row~(c) increases as the K-L divergence in row~(e) increases. This shows that the entropy must increase (from $0$) to ensure some amount of task embodiment.

Synthetic cells can encode these simplified control policies by physically combining their movement, sensory, and memory elements with a series of logical operators, as discussed next, in Section~\ref{sec-control-proj}.

\section{Projecting Policies onto Discrete Sensors} \label{sec-control-proj}

Section~\ref{sec-controls-first} described synthetic cells with perfect state measurement. In this section, implementations using discrete sensors will be explored. In some design processes, it may be possible to \emph{create} sensors that are able to detect exactly where a robot should switch between control modes (e.g., a sensor that can perfectly sense the boundary between the green and orange regions of the control policy). It is also possible that a designer may start with a \emph{fixed} library of sensors, in which case the state space should first be divided into sensed regions and then control modes should be assigned, as described in Section~\ref{sec-sensing-first}. Another possible scenario, and the one we will examine in this section, is that a designer has many sensors to choose from, and will want to use some subset of them.

For the synthetic cell example, we will assume discrete sensing provided by a chemical comparator---a device that compares the relative strength of two chemical concentrations. From a given library of sensors, how should the combination of sensors, actuators, and logical operators be chosen so that the task is best achieved?

\begin{figure}
\centering
\includegraphics[width=7cm]{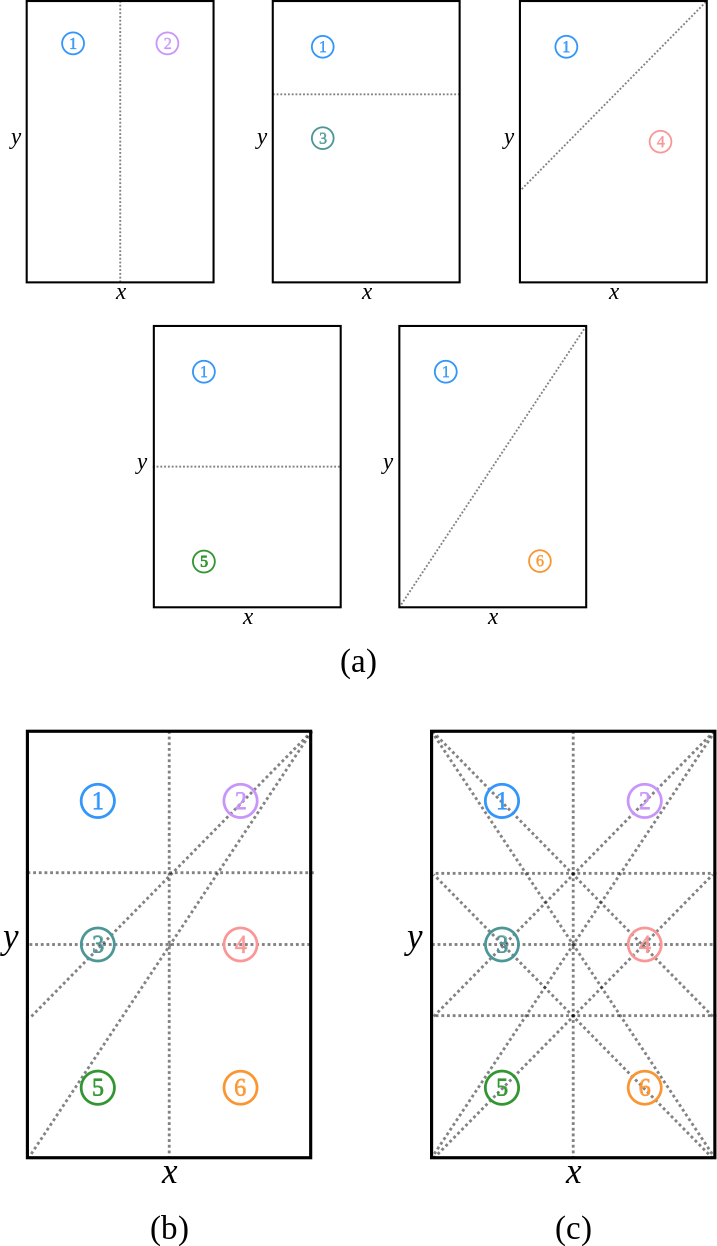}
\caption{(a) Illustration of five individual chemical comparators, each comparing the chemical potential of Source 1 with one of the other sources. Each sensor can tell whether the robot is on one side of an equipotential---the dotted line---or the other. (b) Using a combination of the five sensors, a robot is able to sense which of these $9$ regions it is in. (c) Sensor regions resulting from the combination of all possible chemical comparators in this environment.}
\label{fig-potentials}
\end{figure}

Figure~\ref{fig-potentials}~(a) shows five different individual sensors: each comparing the strength of chemical source 1 to another of the chemical sources in the environment, and how each of these sensors is able to divide the state space into two distinct regions, while Fig.~\ref{fig-potentials}~(b) illustrates which regions of the state space are able to be discerned using these five sensors \emph{combined}. Figure~\ref{fig-potentials}~(c) shows all possible combinations of comparators: all six chemical sources compared to each of their five counterparts, and therefore the maximum granularity of sensed regions in the state space using this sensor library\footnote{Note that some comparators divide the state space in the exact same way, e.g., comparing chemical sources 1 and 3 results in the same sensed regions as comparing sources 2 and 4 (this is true for three other sets of comparators: 1/2 = 3/4 = 5/6, 1/5 = 2/6, and 3/5 = 4/6).}.

The optimal scenario would be that these sensor regions correspond perfectly to the control regions found using the iterative algorithm in Fig.~\ref{fig-policies}. Since this will almost never be the case, we must attempt to approximate our control policy using the library of sensors. 

\begin{figure}
\centering
\includegraphics[width=8cm]{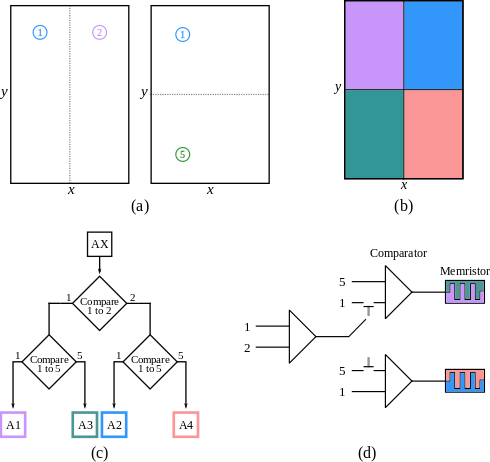}
\caption{Low-fidelity Design. (a) Two sensors. Left: comparing chemical Sources 1 and 2 divides the state space into left and right. Right: comparing Sources 1 and 5 divides the space into top and bottom. (b) Control policy from Fig.~\ref{fig-policies} projected onto the sensed regions.  (c) Logical decision diagram for this system. (d) Circuit diagram for physical synthetic cell design.}
\label{fig-lowfi}
\end{figure}

\begin{figure}
\centering
\includegraphics[width=8cm]{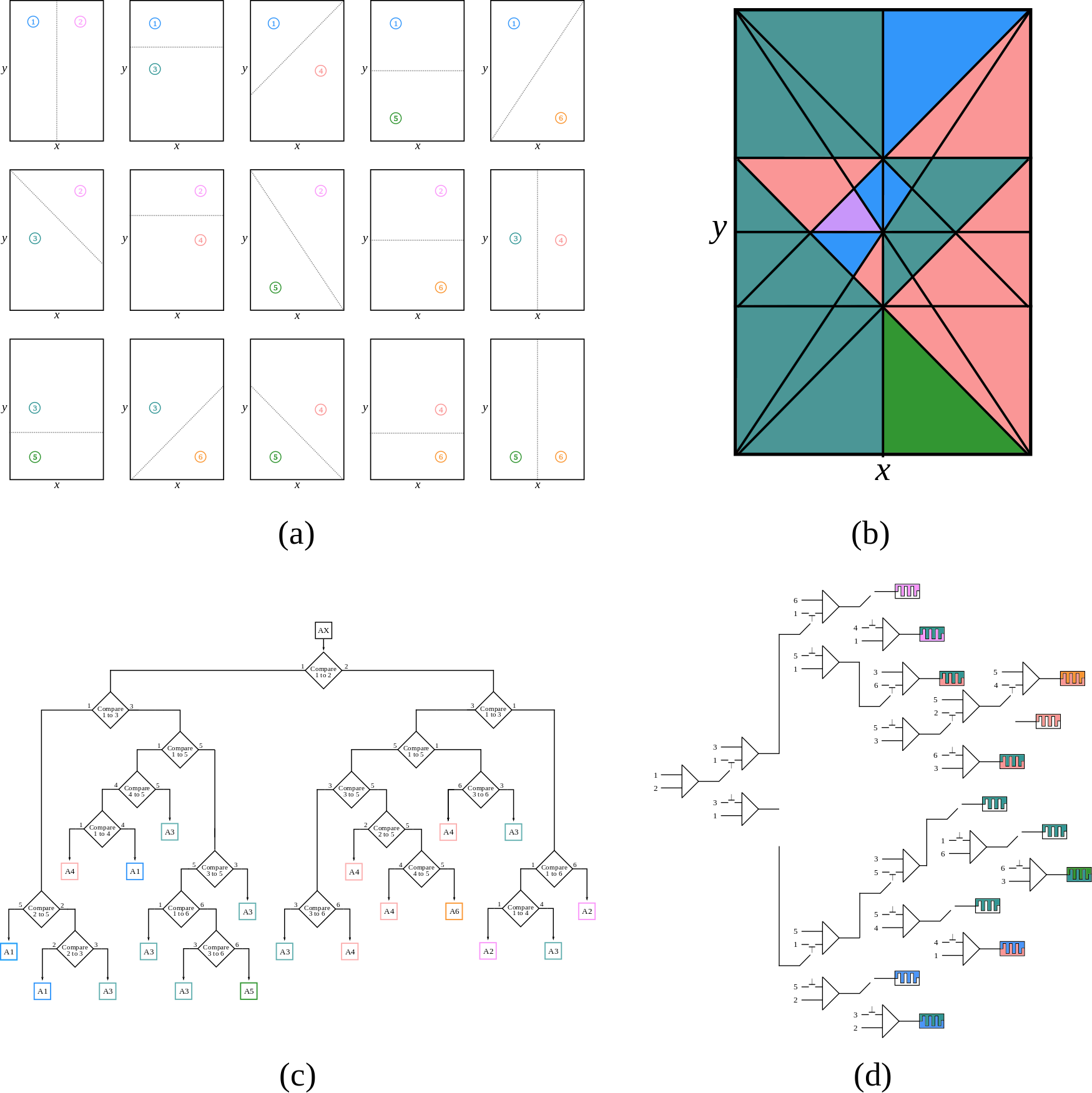}
\caption{High-fidelity Design. (a) Ten distinct sensors. The equipotential lines demonstrate how the device can use chemical comparators to estimate its location in the environment. (b) Control policy from Fig.~\ref{fig-policies} projected onto the sensed regions. (c) Logical decision diagram for this system. (d) Circuit diagram for physical synthetic cell design.}
\label{fig-highfi}
\end{figure}

Figures~\ref{fig-lowfi} and \ref{fig-highfi} demonstrate how the control policy synthesized in the previous section combines with a library of sensors to create a physical design. Figure~\ref{fig-lowfi}~(a) shows two comparators chosen from the sensor library and how they each divide the state space into sensed regions and Fig.~\ref{fig-lowfi}~(b) is the policy that results from projecting the final control policy found in the algorithm onto the feasible sensor space.

This projection from continuous sensing to discrete sensing is done by simulating many rollouts, and finding the best discrete control mode for each sensor region. We pose the question: assuming the best possible control (the unconstrained-sensing control policy from Fig.~\ref{fig-policies}) everywhere else in the state space, which control mode should be used inside each individual sensor region?

In each rollout, a trajectory begins from a random initial condition $x$ within the state space (and therefore a random initial sensor region $r$ in the set of sensor regions $R$). As the simulated synthetic cell executes its trajectory, it uses a single control mode $s$ when it is inside its initial region $r$, and the continuous sensing control policy $C(x)$ (from Fig.~\ref{fig-policies}) when it is outside that initial region $r$. This is executed for each of the $S$ control modes, and the cost $J$ of each trajectory is calculated based on how long it takes the particle to reach the desired point. In case a trajectory gets stuck in a loop, or for some other reason never reaches the desired point, the algorithm will break the loop after $i_{max}$ increments, and record a large cost for that control mode and sensor region combination. After $N$ rollouts, we assign the lowest cost control mode $s$ to each sensor region $r$ to construct the projected, discrete control policy $D(r)$. As $N \rightarrow \infty$, repeated execution of the algorithm will not change the resulting policy $D(r)$. The projection algorithm also does not depend on the order of executions, and can be computed in parallel. An outline of this process can be found in Algorithm~\ref{alg-2}.

\begin{algorithm}
\caption{Projection}
\label{alg-2}
\begin{algorithmic}
\NoDo
\For {each rollout $n \in N$}
    \State $x = $ random initial condition 
    \State $r = $ initial sensor region of $x$
    \State $i = 0$ (increment counter)
    \NoDo
    \For {each control mode $s \in S$}
	    \NoDo
	    \While {$x \neq$ desired point $P$}
	    \NoThen
    		\If {current sensor region of $x = r$}
    		    \State $u = s$
    		\Else 
    		    \State $u = $ control from continuous policy $C(x)$
    		\EndIf
		\State Update $x$ using dynamics $f(x,u)$
		\State Update cost $J$
		\If {$i > i_{max}$}
    		\State Break loop and record large cost
		\EndIf
    	\EndWhile
	\State Record cost $J$ for each region $r$ and control mode $s$
	\EndFor
\EndFor
\State Generate discrete policy $D(r) = \argmin\limits_{s} J(r) \ \ \forall r \in R$
\end{algorithmic}
\end{algorithm}

Logical operators can be combined with sensory observations to represent the state space with more fidelity than sensors alone (e.g., a single sensor in Fig.~\ref{fig-potentials}~(a))---so that actions can be associated with \emph{combinations} of sensory observations (e.g., Fig.~\ref{fig-potentials}~(b)). Figure~\ref{fig-lowfi}~(c) illustrates the logical diagram that would be physically encoded in circuitry onto a synthetic cell so that the policy in Fig.~\ref{fig-lowfi}~(b) could be executed.

Figure~\ref{fig-highfi} is similar to Fig.~\ref{fig-lowfi}, but illustrates the physical design corresponding to the highest fidelity control policy from the library of comparator sensors. Figure~\ref{fig-highfi}~(a) shows each of the sensors in the library, including the ones that repeat sensed regions due to the symmetry in this environment. The projected control policy shown in Fig.~\ref{fig-highfi}~(b) and Fig.~\ref{fig-highfi}~(c) illustrates the logic of the physical circuitry.

\begin{figure*}[h]
\centering
\includegraphics[width=0.9\textwidth]{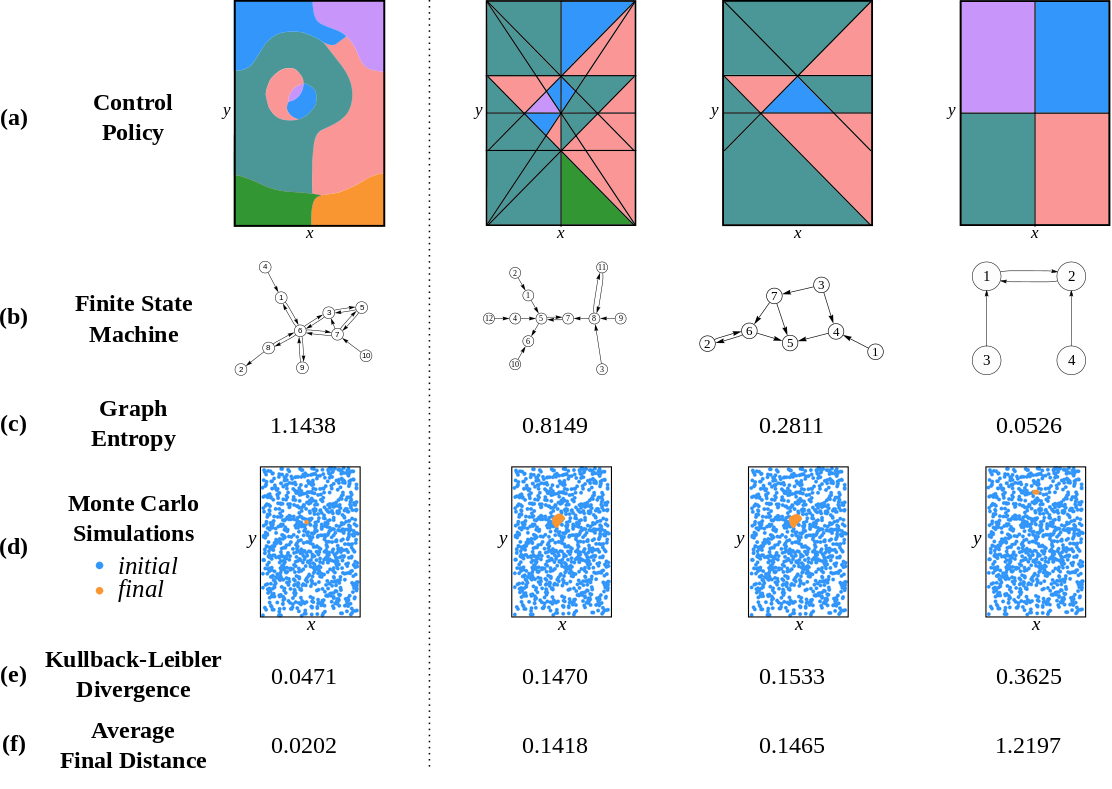}
\caption{Left: The policy generated when perfect knowledge of state was assumed. Right: A high-fidelity design using all of the ($10$ distinct) sensors in the sensor library, as shown in Fig.~\ref{fig-highfi}. A medium-fidelity design, using five sensors. A low-fidelity design, using only two of the sensors from the sensor library, as shown in Fig.~\ref{fig-lowfi}.}
\label{fig-designs}
\end{figure*}

It is notable that the designs in Figures~\ref{fig-lowfi} and \ref{fig-highfi} are quite dissimilar. Figure~\ref{fig-designs} shows how each of the physically feasible designs compare to each other, to another physically feasible design, and to the control policy with perfect knowledge of state. The high-fidelity design in the middle of Fig.~\ref{fig-designs} captures much of the structure of the sensor-agnostic policy, and the results are evident in the relatively low K-L divergence. The medium-fidelity design uses fewer sensors than the high-fidelity one and consequently does not embody the task quite as well. The low-fidelity design has the highest K-L divergence, corresponding to the worst task performance. But, depending on the goals of the designer, it's possible that even this task performance is good enough to sufficiently achieve the goal, and a synthetic cell would be designed in this simplest form. 

\section{Control Policy Generation: Sensing First} \label{sec-sensing-first}

The main objective of this paper has been to encode task information in material properties to produce a simple, physically feasible synthetic cell design that will best achieve a task. We have specified this problem statement to include a set library of sensors (chemical comparators) and actuators (attraction to chemical sources), and in Sections~\ref{sec-controls-first} and \ref{sec-control-proj} we discussed taking a finite set of control modes, finding a control policy using those discrete control modes and assuming perfect sensing, and then projecting that policy onto discrete sensors. But there may be cases where a robot designer has good reason to solve this problem in the opposite order.

In this section, we will discuss generating a control policy with discrete sensors and assuming continuous control authority, and then projecting this policy onto discrete control modes.

\begin{figure}[ht]
\centering
\includegraphics[width=8.4cm]{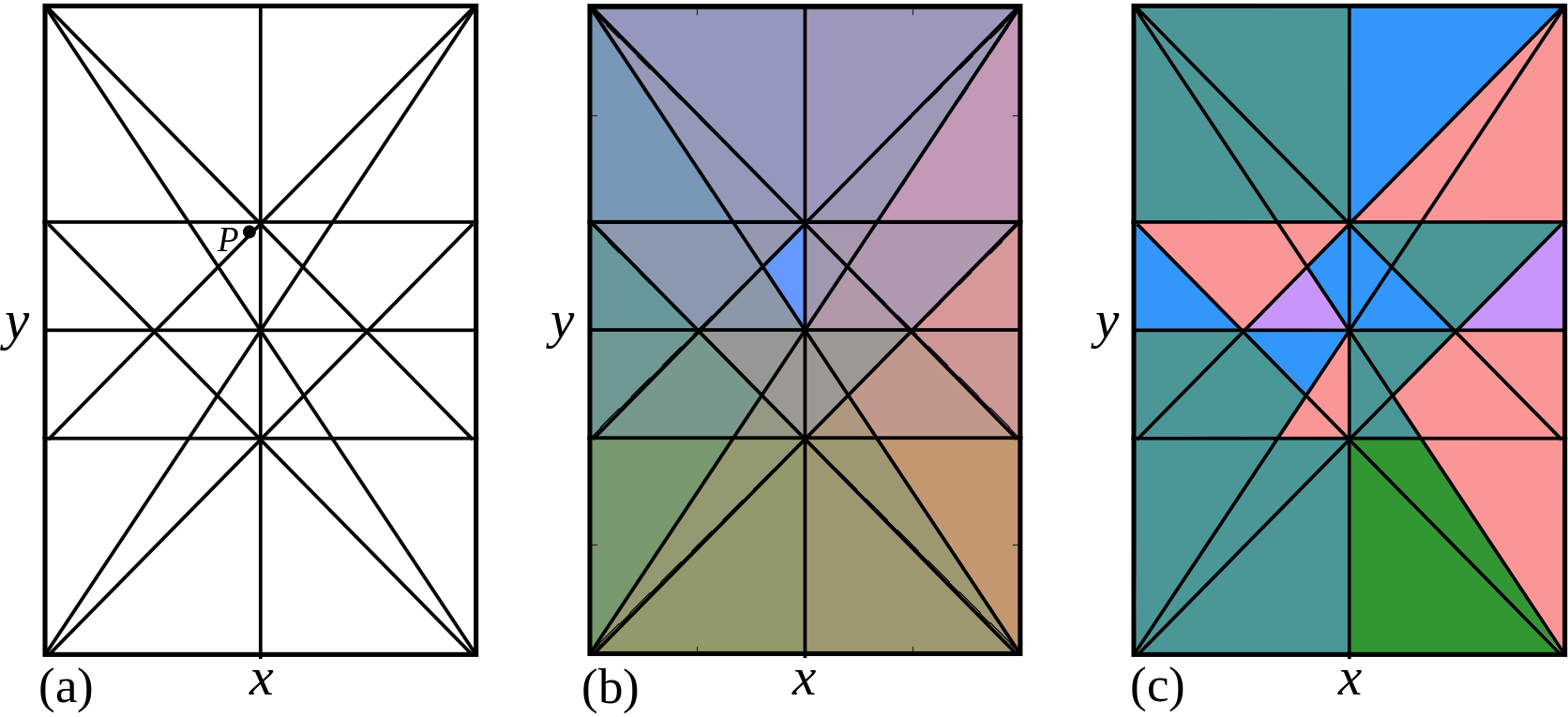}
\caption{(a) Combinations of ten discrete sensors yield 32 distinct regions in the state space. (b) Control policy generated with discrete sensing and continuous control authority (chemical sources placed anywhere in the environment). (c) Control policy projected onto discrete control modes.}
\label{fig-sensor-proj}
\end{figure}

Figure~\ref{fig-sensor-proj}~(a) shows the $(x,  y)$ state space divided into $32$ regions using chemical comparators. To compute a control policy assuming continuous control capabilities, we use the same type of control authority as the previous sections (the synthetic cell being attracted to a chemical potential) but the location of that chemical potential is no longer restricted to a few fixed positions.

Figure~\ref{fig-source-locations} shows an RGB (red, green, blue) color gradient that represents all possible locations of a chemical source in this system. The value of red is increased as the $x$ location of a potential source increases (i.e., moves from left to right) and the value of blue is increased as the position of the chemical potential increases in the $y$ direction (i.e., moves from the bottom to the top). The environment contains a constant amount of green, so where $x$ and $y$ are both small, the color green is most visible (e.g., in the bottom left corner, at $(x=0,y=0)$). This color-based representation of $(x,y)$ locations of potential sources allows us to illustrate continuous control authority. 

A control policy was generated using the discrete sensor regions shown in Fig.~\ref{fig-sensor-proj}~(a) and the continuous control shown on the left of Fig.~\ref{fig-source-locations}. The resulting policy is shown in Fig.~\ref{fig-sensor-proj}~(b). This was computed using rollouts: for each sensor region, a control mode (i.e., a location for a chemical potential to be placed) was chosen that would minimize the cost of trajectories starting in that region, and its location is illustrated by the color of each region. The left of Fig.~\ref{fig-designs2} shows the performance of this policy.

Note that we are using the same sensor regions (divided by equipotential lines from Fig.~\ref{fig-potentials}) as in the last section, even though we are choosing new locations for the chemical potentials. On synthetic cells, the ability to detect and compare specific chemicals (as a chemical comparator does) and the ability to be attracted to a certain chemical (resulting in locomotion) are distinct and unrelated. There might be many chemical stimuli in an environment that a synthetic cell can use for state estimation which have no affect at all on a cell's actuation. Here we continue with the assumption that the synthetic cell has chemical comparators on board that pertain to the $6$ potentials shown in Fig.~\ref{fig-control-proj}, but assert that there is some possibility in the design space that sources might be added at new positions, or that we otherwise would be interested to know where the best possible source locations are.

\section{Projecting Policies Onto Discrete Actuators} \label{sec-sensing-proj}

Now that we have calculated our control policy with discrete sensors and unconstrained control authority, the optimal scenario would be to create actuators that perfectly align with the policy: if possible, we should place a chemical potential at each preferred location. Since this is unlikely to be easily achievable, we will approximate our control policy using a library of actuators.

\begin{figure}[t]
\centering
\includegraphics[width=0.48\textwidth]{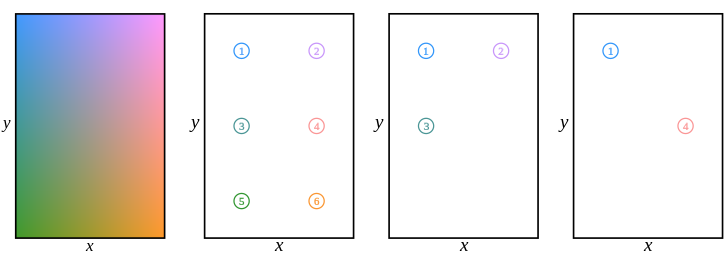}
\caption{Left: All possible locations of chemical sources, represented by RGB (red, blue, green) colors. As the $x$ location of a source increases, the color illustrating the source location becomes more red. As the $y$ location of the source increases, more blue is added to the color. There is a constant level of the color green throughout the state space. Right: Three different subsets of discrete control modes are shown: $6$ discrete source locations, $3$ locations, and only $2$ locations.}
\label{fig-source-locations}
\end{figure}

\begin{figure*}
\centering
\includegraphics[width=0.9\textwidth]{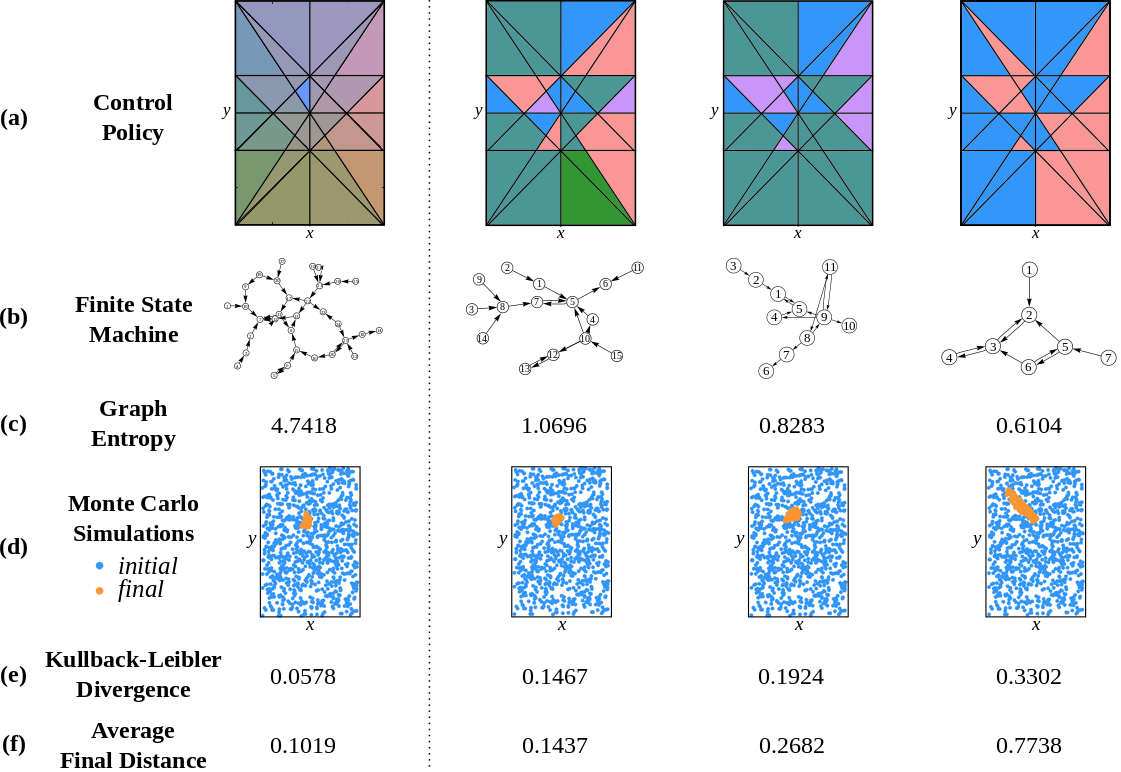}
\caption{Left: The policy generated when continuous control was assumed. Right: A high-controllability design using all $6$ of the discrete control modes in the actuator library, as shown in Fig.~\ref{fig-source-locations}. A medium-controllability design using only three control modes (chemical sources $1$, $2$, and $3$). A low-controllability design using only two of the actuators from the actuator library: chemical sources $1$ and $4$.}
\label{fig-designs2}
\end{figure*}

For continuity, we assume that our library of actuators consists of the same control modes shown in Fig.~\ref{fig-control-proj}. We project the computed policy onto 3 subsets of these actuators, shown on the right side of Fig.~\ref{fig-source-locations}. The subsets are: all six control modes, a set of three of the control modes ($1$, $2$, and $3$) and a set of only two control modes ($1$ and $4$). We generate synthetic cell designs with each of these sets of actuators by projecting the unconstrained controls computed in the policy to these discrete control modes, and then compare their design complexities and levels of task embodiment in Fig.~\ref{fig-designs2}.

The projection operator in this section is the same as the one described in Sec.~\ref{sec-control-proj}. But in this case, the continuous control policy $C(x)$ (used outside each region being evaluated) is the discrete-sensing, continuous-actuation policy shown in Fig.~\ref{fig-sensor-proj}, rather than the discrete-actuation, continuous sensing control policy shown in Fig.~\ref{fig-control-proj}.

Figure~\ref{fig-designs2} shows the projected control policies and the results of simulations performed with each different design. Monte Carlo simulations were performed with random initial conditions, shown in row (d), and the average distance of the final points from the desired point is shown in row (f). Note the trends in rows (c) and (e). Similarly to Fig.~\ref{fig-designs}, we observe that as the graph entropy in row (c) decreases, the K-L divergence in row (e) decreases. This shows that as the designs become simpler (i.e., use fewer control modes), the task performance becomes worse. But, depending on the goals of the robot design process, it's possible that even the worst performance shown on the right would be a worthy trade off for the simplicity of the design and ease of fabrication.

\section{Discussion} \label{sec-disc}
In this work we addressed the question of designing robots while minimizing complexity and maximizing task embodiment. We demonstrated our method of solving this min-max problem, which included both solving for the organization of actuators first and then projecting onto discrete sensors, and organizing sensors first and then projecting onto discrete actuators. To accomplish the former, an iterative algorithm was introduced that resulted in a simple control policy assuming discrete control modes and perfect sensing, and then projecting that policy onto a discrete space of sensed regions resulting from a library of sensors. This is not necessarily an optimal design pipeline for all robot design problems.  
There may be some instances where there is a fixed library of sensors, in which case one would solve the latter problem, by first dividing the state space into discrete sensor regions, then computing a control policy assuming discrete sensors and unconstrained control authority, and finally projecting the policy onto a discrete library of control modes from a library of actuators.

When these two approaches were applied to the same libraries of sensors and actuators, slightly different designs were generated (as seen in the second columns of Fig.~\ref{fig-designs} and Fig.~\ref{fig-designs2}). Also, although both methods resulted in similar levels of task embodiment across the different designs, the designs produced by the sensors-first method were more complex. This is because the complexity of a design is directly related to the number of states, which depends significantly on the quantity of sensor regions. In Fig.~\ref{fig-control-proj} each subsequent design had fewer and fewer sensors where as in Fig.~\ref{fig-sensor-proj} each design had the same number of sensor regions, keeping the complexity relatively high.

In future work, this methodology will be implemented for a wider range of dynamical systems, specifically higher order systems. The algorithm will be tested with different modifications, including using $D_{KL}$ in the objective function so that task embodiment is the actual object of the optimization, rather than a correlated consequence. The authors also plan to use reinforcement learning techniques to organize sensors and actuators simultaneously, over many rollouts. Finally, this methodology will be validated by using control policies computed by this method to design and create actual, physical synthetic cells.

\section*{Acknowledgment}
The authors would like to thank Dr. Michael Strano and Albert Tianxiang Liu who provided insight and expertise that greatly assisted this research.


\begin{IEEEbiography}[{\includegraphics[width=1in,height=1.25in,clip,keepaspectratio]{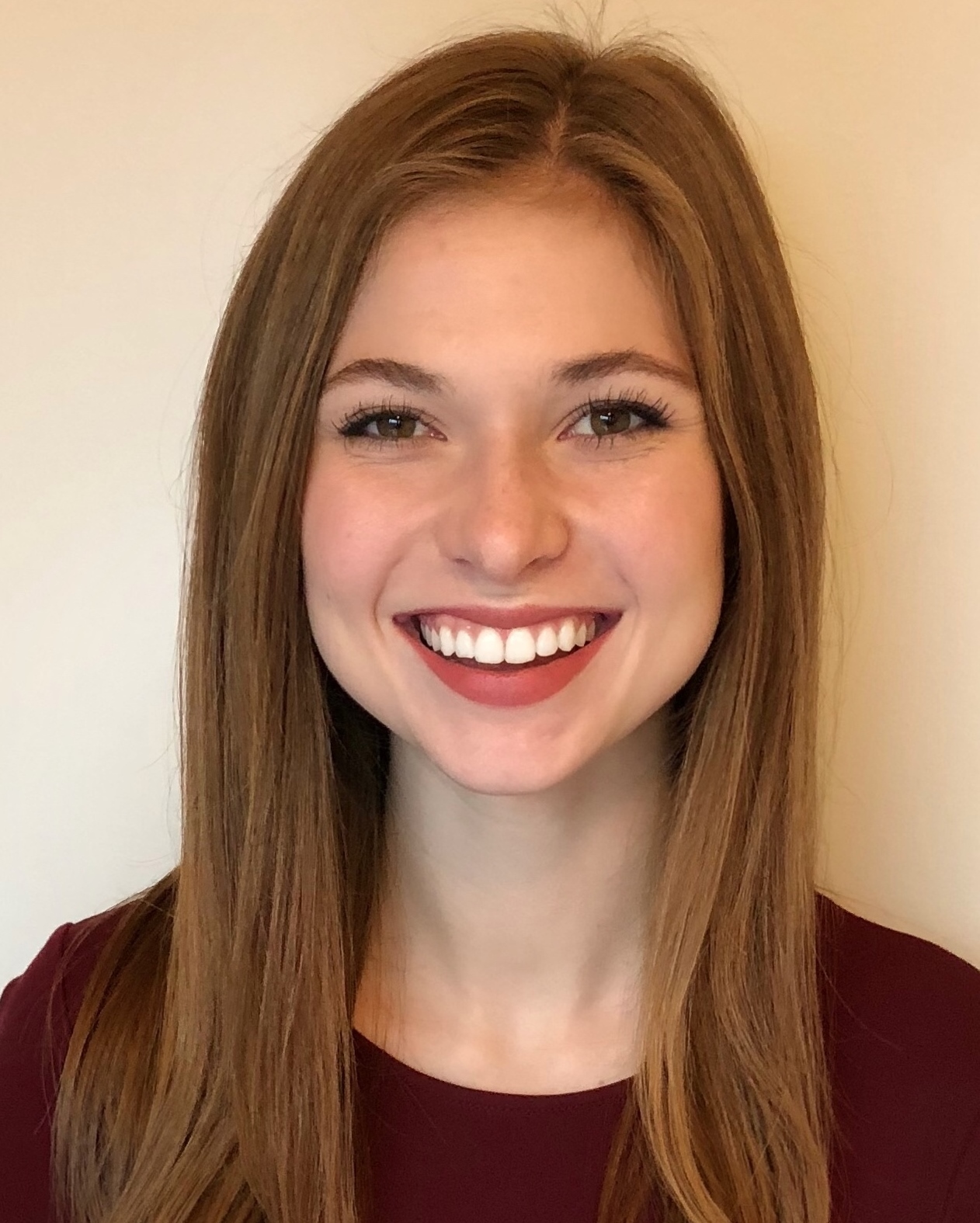}}]{Ana Pervan}
Ana Pervan received her B.S. degree in Mechanical Engineering from the University of Notre Dame in 2016 and her M.S. degree in Mechanical Engineering from Northwestern University in 2018. She is currently a PhD candidate in Mechanical Engineering at Northwestern University. Her research interests are focused in algorithmic design of robots and emergent behaviors. She was awarded the National Science Foundation Graduate Research Fellowship in 2018.
\end{IEEEbiography}

\begin{IEEEbiography}[{\includegraphics[width=1in,height=1.25in,clip,keepaspectratio]{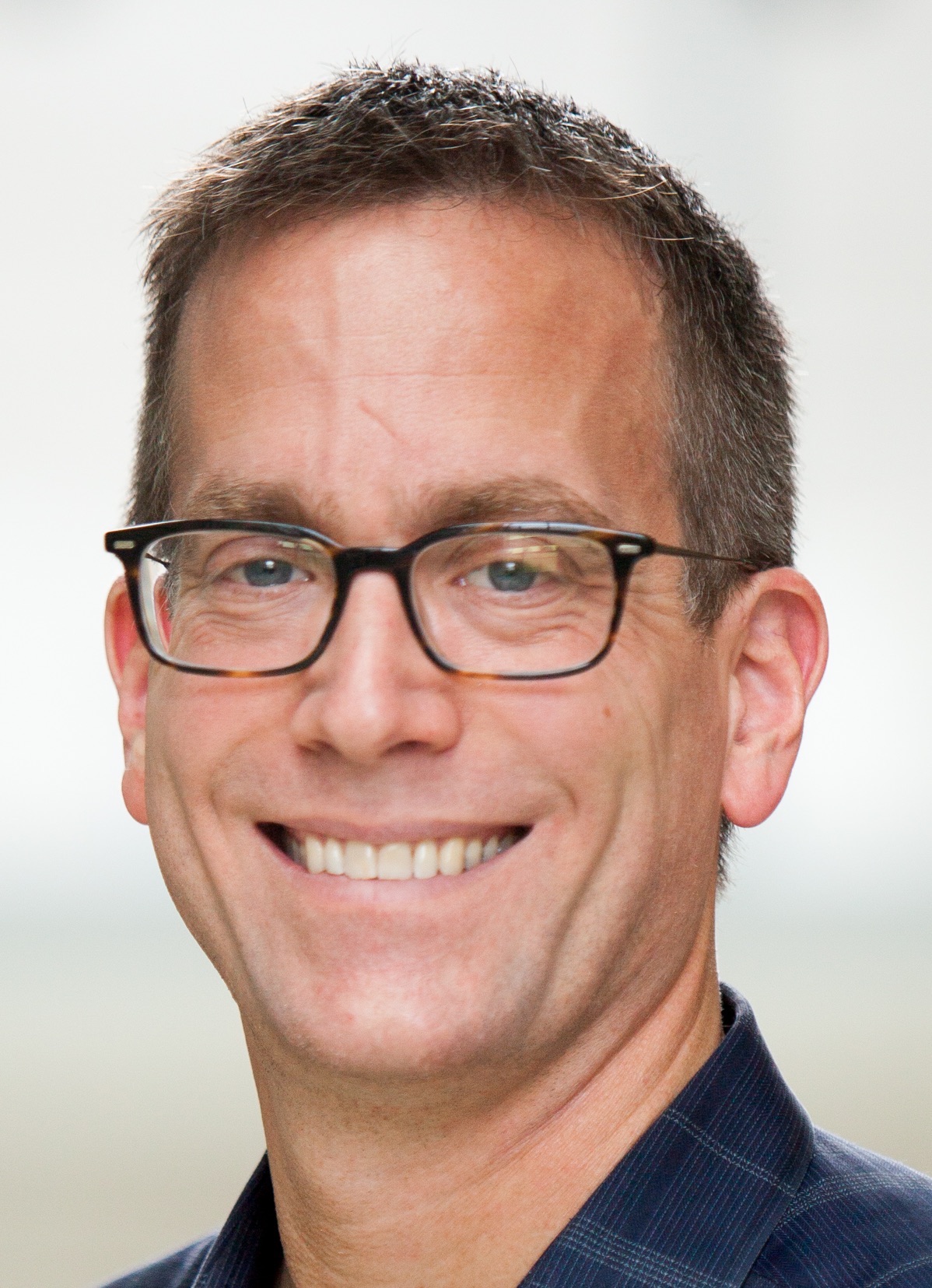}}]{Todd D. Murphey}
Todd D. Murphey received his B.S. degree in mathematics from the University of Arizona and the Ph.D. degree in Control and Dynamical Systems from the California Institute of Technology.
He is a Professor of Mechanical Engineering at Northwestern University. His laboratory is part of the Neuroscience and Robotics Laboratory, and his research interests include robotics, control, computational methods for biomechanical systems, and computational neuroscience.
Honors include the National Science Foundation CAREER award in 2006, membership in the 2014-2015 DARPA/IDA Defense Science Study Group, and Northwestern’s Professorship of Teaching
Excellence. He was a Senior Editor of the IEEE Transactions on Robotics.
\end{IEEEbiography}
\vfill





\begin{thebibliography}{10}
\small
\bibitem{anand2009entropy}
K.~Anand and G.~Bianconi.
\newblock Entropy measures for networks: Toward an information theory of
  complex topologies.
\newblock {\em Physical Review E}, Oct 2009.

\bibitem{armijo1966}
L.~Armijo.
\newblock Minimization of functions having lipschitz continuous first partial
  derivatives.
\newblock {\em Pacific Journal of Mathematics}, pages 1--3, 1966.

\bibitem{banarse2019}
D.~Banarse, Y.~Bachrach, S.~Liu, G.~Lever, N.~Heess, C.~
Fernando, P.~Kohli, and T.~Graepel.
\newblock The Body is Not a Given: Joint Agent Policy Learning and Morphology Evolution.
\newblock {\em International Conference on Autonomous Agents and Multiagent Systems
(AAMAS)}, May 2019.

\bibitem{belta2007symbolic}
C.~Belta, A.~Bicchi, M.~Egerstedt, E.~Frazzoli, E.~Klavins, and G.~J. Pappas.
\newblock Symbolic planning and control of robot motion [grand challenges of
  robotics].
\newblock {\em IEEE Robotics Automation Magazine}, pages 61--70, Mar 2007.

\bibitem{decarlo}
S.~C. Bengea and R.~A. DeCarlo.
\newblock Optimal control of switching systems.
\newblock {\em Automatica}, pages 11--27, Jan 2005.

\bibitem{bishop2006}
C.~M. Bishop.
\newblock {\em Pattern Recognition and Machine Learning}.
\newblock Springer, 2006.

\bibitem{brooks1991}
R.~A.~Brooks.
\newblock New approaches to robotics.
\newblock {\em Science}, Sept 1991.

\bibitem{burgess2017}
C.~Burgess, P.~Hayman, M.~Williams, et al.
\newblock Point-particle effective field theory I: classical renormalization and the inverse-square potential.
\newblock {\em J. High Energ. Phys.}, Apr 2017.

\bibitem{caldwell2013projection}
T.~M. Caldwell and T.~D. Murphey.
\newblock Projection-based optimal mode scheduling.
\newblock {\em IEEE Conference on Decision and Control}, Dec 2013.

\bibitem{caldwellNAHS}
T.~M. Caldwell and T.~D. Murphey.
\newblock Projection-based iterative mode scheduling for switched systems.
\newblock {\em Nonlinear Analysis: Hybrid Systems}, pages 59--83, Aug 2016.

\bibitem{censi2015codesign}
A.~Censi.
\newblock A mathematical theory of co-design.
\newblock {\em ArXiv e-prints https://arxiv.org/abs/1512.08055}, 2015.

\bibitem{censi2016cyclic}
A.~Censi.
\newblock A Class of Co-Design Problems With Cyclic Constraints and Their Solution.
\newblock {\em IEEE Robotics and Automation Letters}, Jan 2016.

\bibitem{censi2017uncertainty}
A.~Censi.
\newblock Uncertainty in Monotone Co-Design Problems.
\newblock{\em IEEE Robotics and Automation Letters}, Feb 2017.

%
\bibitem{collins2005bipedal}
S.~Collins, A.~Ruina, R.~Tedrake, and M.~Wisse.
\newblock Efficient bipedal robots based on passive-dynamic walkers.
\newblock {\em Science}, pages 1082--1085, 2005.

\bibitem{dehmer2011graphentropy}
M.~Dehmer and A.~Mowshowitz.
\newblock A history of graph entropy measures.
\newblock {\em Journal of Information Science}, pages 57--78, Jan 2011.

\bibitem{dickinson2000}
M.~H.~Dickinson, C.~T.~Farley, R.~J.~Full, M.~A.~R.~Koehl, R.~Kram, S.~Lehman.
\newblock How Animals Move: An Integrative View.
\newblock {\em Science}, pages 100--106, Apr 2000.

\bibitem{egerstedt2006transitiontime}
M.~Egerstedt, Y.~Wardi, and H.~Axelsson.
\newblock Transition-time optimization for switched-mode dynamical systems.
\newblock {\em IEEE Transactions on Automatic Control}, pages 110--115, Jan 2006.

\bibitem{Egerstedt03optimalcontrol}
M.~Egerstedt, Y.~Wardi, and H.~Axelsson.
\newblock Optimal control of switching times in hybrid systems.
\newblock {\em International Conference on Methods and Models in Automation and Robotics}, 2003.

\bibitem{erdmann1995}
M.~Erdmann.
\newblock Understanding action and sensing by designing action-based sensors.
\newblock {\em The International Journal of Robotics Research}, pages 483--509, 1995.
%
\bibitem{esteban2018nano}
B.~Esteban-Fern{\'a}ndez~de {\'A}vila, P.~Angsantikul, D.~E.
  Ram{\'\i}rez-Herrera, F.~Soto, H.~Teymourian, D.~Dehaini, Y.~Chen, L.~Zhang, and J.~Wang.
\newblock Hybrid biomembrane functionalized nanorobots for concurrent removal
  of pathogenic bacteria and toxins.
\newblock {\em Science Robotics}, 2018.

\bibitem{howard2019}
D.~Howard, A.~E.~Eiben, D.~F.~Kennedy, J.~B.~Mouret, P.~Valencia and D.~Winkler.
\newblock   Evolving embodied intelligence from materials to machines.
\newblock {\em Nature Machine Intelligence}, 2019.

\bibitem{laviers}
U.~Huzaifa, C.~Bernier, Z.~Calhoun, G.~Heddy, C.~Kohout, B.~Libowitz,
  A.~Moenning, J.~Ye, C.~Maguire, and A.~LaViers.
\newblock Embodied movement strategies for development of a core-located
  actuation walker.
\newblock {\em 2016 6th IEEE International Conference on Biomedical Robotics
  and Biomechatronics (BioRob)},  pages 176--181, June 2016.

\bibitem{karaman2011sampling}
S.~Karaman and E.~Frazzoli.
\newblock Sampling-based algorithms for optimal motion planning.
\newblock {\em The International Journal of Robotics Research}, pages 846--894, 2011.


%
\bibitem{li2018nano}
S.~Li, Q.~Jiang, S.~Liu, Y.~Zhang, Y.~Tian, C.~Song, J.~Wang, Y.~Zou, G.~J. Anderson, J.~Han, Y.~Chang, Y~.Liu, 
C.~Zhang, L.~Chen, G.~Zhou, G.~Nie, H.~Yan, B.~Ding, and Y.~Zhao.
\newblock A DNA nanorobot functions as a cancer therapeutic in response to a
  molecular trigger in vivo.
\newblock {\em Nature Biotechnology}, pages 258--264, 2018.

\bibitem{janus}
P.~Liu, A.~T. Liu, D.~Kozawa, J.~Dong, J.~F. Yang, V.~B. Koman, M.~Saccone, S.~Wang, Y.~Son, M.~H.~Wong, and M.~S. Strano 
\newblock Autoperforation of 2D materials for generating two-terminal memristive Janus particles.
\newblock {\em Nature Materials}, pages 1005--1012, 2018.

\bibitem{strano}
P.~Liu, A.~L. Cottrill, D.~Kozawa, V.~B. Koman, D.~Parviz, A.~T. Liu, J.~F. Yang, T.~Q. Tran, M.~H. Wong, S.~Wang, and M.~S. Strano.
\newblock Emerging trends in 2D nanotechnology that are redefining our
  understanding of “nanocomposites”.
\newblock {\em Nano Today}, 2018.

\bibitem{mavrommati2016cap}
A.~Mavrommati and T.~D. Murphey.
\newblock Automatic synthesis of control alphabet policies.
\newblock {\em IEEE International Conference on Automation Science and
  Engineering (CASE)}, Aug 2016.
  
%
\bibitem{arelekatti201prosthetic}
V.~N. Murthy~Arelekatti and A.~G. Winter, V.
\newblock Design and preliminary field validation of a fully passive prosthetic
  knee mechanism for users with transfemoral amputation in india.
\newblock {\em Journal of Mechanisms and Robotics}, pages 350--356, 2015.

\bibitem{okane2008power}
J.~M. O’Kane and S.~M. LaValle.
\newblock Comparing the power of robots.
\newblock {\em The International Journal of Robotics Research}, pages 5--23, Jan 2008.

\bibitem{okane2017concise}
J.~M. O’Kane and D.~A. Shell.
\newblock Concise Planning and Filtering: Hardness and Algorithms.
\newblock {\em IEEE Transactions on Automation Science and Engineering}, pages 1666--1681, Oct 2017.

\bibitem{pervan2018wafr}
A.~Pervan and T.~D.~Murphey.
\newblock Low complexity control policy synthesis for cyber-free robot design.
\newblock {\em Workshop on the Algorithmic Foundations of Robotics (WAFR)}, Dec 2018.


\bibitem{pfeifer2007book}
R.~Pfeifer, J.~C.~Bongard.
\newblock {\em How the Body Shapes the Way We Think}.
\newblock MIT Press, 2007.

\bibitem{pfeifer2007science}
R.~Pfeifer, M.~Lungarella, F.~Iida.
\newblock Self-organization, embodiment, and biologically inspired robotics.
\newblock {\em Science}, pages 1088--1093, Nov 2007.

\bibitem{pola}
G.~Pola, A.~Girard, and P.~Tabuada.
\newblock Approximately bisimilar symbolic models for nonlinear control
  systems.
\newblock {\em Automatica}, pages 2508--2516, 2008.

\bibitem{ramadage1987}
P.~J. Ramadge and W.~M. Wonham.
\newblock Supervisory control of a class of discrete event processes.
\newblock {\em SIAM Journal on Control and Optimization}, pages 206--230, 1987.

\bibitem{shaikh2003optimal}
M.~Shahid Shaikh and P.~E. Caines.
\newblock On the optimal control of hybrid systems: Optimization of
  trajectories, switching times, and location schedules.
\newblock  {\em Hybrid Systems: Computation and Control}. Springer Berlin Heidelberg, 2003.

\bibitem{spagna2007}
J.~C.~Spagna, D.~I.~Goldman, P.~C.~Lin, D.~E.~Koditschek, and R.~J.~Full.
\newblock Distributed mechanical feedback in arthropods and robots simplifies control of rapid running on challenging terrain.
\newblock  {\em Bioinspiration and Biomimetics}, Jan 2007.

%
\bibitem{tedrake2004walker}
R.~Tedrake, T.~W. Zhang, M.~Fong, and H.~S. Seung.
\newblock Actuating a simple 3D passive dynamic walker.
\newblock {\em IEEE International Conference on Robotics and Automation}, Apr 2004.

\bibitem{shell}
J.~Wainer, D.~J. Feil-Seifer, D.~A. Shell, and M.~J. Mataric.
\newblock Embodiment and human-robot interaction: A task-based perspective.
\newblock {\em IEEE International Symposium on Robot
  and Human Interactive Communication}, Aug 2007.

\bibitem{wood2008}
R.~J.~Wood.
\newblock The First Takeoff of a Biologically Inspired At-Scale Robotic Insect.
\newblock {\em IEEE Transactions on Robotics}, 2008.

\bibitem{xu2002optimal}
X.~Xu and P.~J. Antsaklis.
\newblock Optimal control of switched systems via non-linear optimization based
  on direct differentiations of value functions.
\newblock {\em International Journal of Control}, pages 1406--1426, 2002.

\end{thebibliography}
\end{document}